\documentclass{article}

\usepackage{natbib}

\usepackage{url}            
\usepackage{booktabs}       
\usepackage{amsfonts}       
\usepackage{graphicx} 

\usepackage{float}
\usepackage{amsmath}
\usepackage{amssymb}
\usepackage{amsthm}

\floatstyle{ruled}
\newfloat{algorithm}{tbp}{loa}
\providecommand{\algorithmname}{Algorithm}
\floatname{algorithm}{\protect\algorithmname}

\theoremstyle{plain}
\newtheorem{thm}{\protect\theoremname}
\theoremstyle{definition}
\newtheorem{defn}[thm]{\protect\definitionname}
\theoremstyle{plain}
\newtheorem{lem}[thm]{\protect\lemmaname}
\theoremstyle{plain}

\ifx\proof\undefined
\newenvironment{proof}[1][\protect\proofname]{\par
\normalfont\topsep6\p@\@plus6\p@\relax
\trivlist
\itemindent\parindent
\item[\hskip\labelsep\scshape #1]\ignorespaces
}{%
\endtrivlist\@endpefalse
}
\providecommand{\proofname}{Proof}
\fi

\usepackage{algorithmic}

\providecommand{\corollaryname}{Corollary}
\providecommand{\definitionname}{Definition}
\providecommand{\lemmaname}{Lemma}
\providecommand{\theoremname}{Theorem}

\usepackage{enumitem}
\setlist[itemize]{noitemsep}

\usepackage{booktabs}

\allowdisplaybreaks

\usepackage{authblk}
\usepackage[bookmarks=false]{hyperref}       

\title{The Second Order Linear Model}

\author[1]{Ming Lin\thanks{linmin@umich.edu}}
\author[1]{Shuang Qiu\thanks{qiush@umich.edu}}
\author[2]{Bin Hong\thanks{bin\_hong@zju.edu.cn}}
\author[1]{Jieping Ye\thanks{jpye@umich.edu}}
\affil[1]{Department of Computational Medicine and Bioinformatic\\
  University of Michigan, Ann Arbor, MI 48109}
\affil[2]{State Key Lab of CAD\&G\\
	Zhejiang University, Hangzhou, China, 310058}

\date{\today}

\begin{document}

\maketitle

\begin{abstract}
We study a fundamental class of regression models called the second
order linear model (SLM). The SLM extends the linear model to high
order functional space and has attracted considerable research interest
recently. Yet how to efficiently learn the SLM under full generality
using nonconvex solver still remains an open question due to several
fundamental limitations of the conventional gradient descent learning
framework. In this study, we try to attack this problem from a gradient-free
approach which we call the moment-estimation-sequence (MES) method.
We show that the conventional gradient descent heuristic is biased
by the skewness of the distribution therefore is no longer the best
practice of learning the SLM. Based on the MES framework, we design
a nonconvex alternating iteration process to train a $d$-dimension
rank-$k$ SLM within $O(kd)$ memory and one-pass of the dataset.
The proposed method converges globally and linearly, achieves $\epsilon$
recovery error after retrieving $O[k^{2}d\cdot\mathrm{polylog}(kd/\epsilon)]$
samples. Furthermore, our theoretical analysis reveals that not all
SLMs can be learned on every sub-gaussian distribution. When the instances
are sampled from a so-called $\tau$-MIP distribution, the SLM can
be learned by $O(p/\tau^{2})$ samples where $p$ and $\tau$ are
positive constants depending on the skewness and kurtosis of the distribution.
For non-MIP distribution, an addition diagonal-free oracle is necessary
and sufficient to guarantee the learnability of the SLM. Numerical
simulations verify the sharpness of our bounds on the sampling complexity
and the linear convergence rate of our algorithm.
\end{abstract}

\section{Introduction}

The second order linear model (SLM) is a fundamental class of regression
models and attracts considerable research interest recently. Given
an instance $\boldsymbol{x}\in\mathbb{R}^{d}$, the SLM assumes that
the label $y\in\mathbb{R}$ of $\boldsymbol{x}$ is generated by
\begin{equation}
y=\boldsymbol{x}{}^{\top}\boldsymbol{w}^{*}+\boldsymbol{x}{}^{\top}M^{*}\boldsymbol{x}+\xi\label{eq:second-order-linear-mode}
\end{equation}
where $\{\boldsymbol{w}^{*},M^{*}\}$ are the first order and the
second order coefficients respectively. The $\xi$ is an additive
sub-gaussian noise term. The SLM defined in Eq. (\ref{eq:second-order-linear-mode})
covers several important models in machine learning and signal processing
problems. When $\boldsymbol{w}^{*}=0$ and $M^{*}$ is a rank-one
symmetric matrix, Eq. (\ref{eq:second-order-linear-mode}) is known
as the phase retrieval problem \citep{candes_phase_2011}. While $M^{*}$
is a rank-$k$ symmetric matrix, Eq. (\ref{eq:second-order-linear-mode})
is equal to the symmetric rank-one matrix sensing problem \citep{kueng_low_2014,cai_rop:_2015, yurtsever_sketchy_2017}.
For $\boldsymbol{w}^{*}\not=\boldsymbol{0}$ and $M_{i,j}^{*}=v_{i}v_{j}$
at $i\not=j$ otherwise $M_{i,i}^{*}=0$, Eq. (\ref{eq:second-order-linear-mode})
is called the Factorization Machine (FM) \citep{rendle_factorization_2010}.
When $M_{i,i}^{*}$ is allowed to be non-zero, the model is called
the Generalized Factorization Machine (gFM) \citep{ming_lin_non-convex_2016}
. It is possible to further extend the SLM to high order functional
space which leads to the Polynomial Network model \citep{blondel_polynomial_2016}.
\citet{lin_multi-task_2016} employ the SLM to capture the feature
interaction in multi-task learning.

Although the SLM have been applied in various learning problems, there
is rare research of SLM under a general setting in form of Eq. (\ref{eq:second-order-linear-mode}).
A naive analysis directly following the sampling complexity of the
linear model would suggest $O(d^{2})$ samples in order to learn the
SLM. For high dimensional problems this is too expensive to be useful.
We need a more efficient solution with sampling complexity much less
than $O(d^{2})$, especially when $M^{*}$ is low-rank. This seemingly
simple problem is still an open question by the time of writing this
paper. There are several fundamental challenges of learning the SLM.
Indeed, even the symmetric rank-one matrix sensing problem, a special
case of the SLM, is proven to be hard. Until very recently, \citet{cai_rop:_2015}
partially answered the sampling complexity of this special case on
well-bounded sub-gaussian distribution using trace norm convex programming
under the $\ell_{2}/\ell_{1}$-RIP condition. \citet{yurtsever_sketchy_2017}
develop a conditional gradient descent solver for the symmetric matrix
sensing problem. However, it is still unclear how to solve this special
case using more efficient nonconvex alternating iteration on general
sub-gaussian distribution such as the Bernoulli distribution. Perhaps
the most state-of-the-art research on the SLM is the preliminary work
by \citet{ming_lin_non-convex_2016}. Their result is still weak because
they rely on the rotation invariance of the Gaussian distribution
therefore their analysis cannot be generalized to non-Gaussian distributions.
Their sampling complexity is $O(k^{3}d)$ which is suboptimal compared
to our $O(k^{2}d)$ bound. The readers familiar with convex geometry
might recall the general convex programming method for structured
signal recovery developed by \citet{tropp_convex_2014}. It is difficult
to apply Tropp's method here because it is unclear how to lower-bound
the conic singular value on the decent cone of the SLM. We would like
to refer the above papers for more historical developments on the
related research topics.

In this work, we try to attack this problem from a nonconvex gradient-free
approach which we call the moment-estimation-sequence method. The
method is based on nonconvex alternating iteration in one-pass of
the data stream within $O(kd)$ memory. The proposed method converges
globally and linearly. It achieves $\epsilon$ recovery error after
retrieving $O[k^{2}dp/\tau^{2}\cdot\mathrm{polylog}(kd/\epsilon)]$
samples where $p$ is a constant depending on the skewness and kurtosis
of the distribution and $\tau$ is the Moment Invertible Property
(MIP) constant (see Definition \ref{def:Moment-Invertible-Property}).
When the instance distribution is not $\tau$-MIP, our theoretical
analysis reveals that an addition diagonal-free oracle of $M^{*}$
is necessary and sufficient to guarantee the recovery of the SLM.

The most remarkable trait of our approach is its gradient-free nature.
In nonconvex optimization, the gradient descent heuristic usually
works well. For most conventional (first order) matrix estimation
problems, the gradient descent heuristic happens to be provable \citep{zhao_nonconvex_2015-1}.
In our language, the gradient iteration on these first order problems
happens to form a moment estimation sequence. When training the SLM
on skewed sub-gaussian distributions, the gradient descent heuristic
no longer generates such sequence. The gradient of the SLM will be
biased by the skewness of the distribution which can even dominate
the gradient norm. This bias must be eliminated which motivates our
moment-estimation-sequence construction. Please see subsection \ref{sub:Learning-Nonconvex-Model}
for an in-depth discussion. 

\textbf{Contribution} We present the first provable nonconvex algorithm
for learning the second order linear model. We shows that the SLM
cannot be efficiently learned with naive alternating gradient descent.
We develop a novel technique called the Moment-Estimation-Sequence
method to overcome this difficulty. The presented analysis provides
the strongest learning guarantees published so far by the time of
writing this paper. Particularly, our work provides the first nonconvex
solver for the symmetric matrix sensing and Factorization Machines
on sub-gaussian distribution with nearly optimal sampling complexity.

The remainder of this paper is organized as follows. In Section 2
we introduce necessary notation and background of the SLM. Subsection
2.1 is devoted to the gradient-free learning principle and the MIP
condition. We propose the moment-estimation-sequence method in Section
3. Theorem \ref{thm:global-convergence-rate-of-Moment-Estimation-Sequence}
bounds the convergence rate of our main Algorithm \ref{alg:moment-estimation-sequence-method}.
A sketched theoretical analysis is briefed in Section 4. Our key theoretical
result is Theorem \ref{thm:subgaussian-shifted-CI-RIP} which is the
counterpart of sub-gaussian Hanson-Wright inequality \citep{rudelson_hanson-wright_2013}
on low-rank matrix manifold. Section 5 conducts numerical simulations
to verify our theoretical results. Section 6 concludes this work.

\section{Notation and Background}

Suppose we are given $n$ training instances $\boldsymbol{x}^{(i)}$
for $i\in\{1,\cdots,n\}$ and the corresponding labels $y_{i}$ identically
and independently (i.i.d.) sampled from a joint distribution $P(\boldsymbol{x},y)$.
Denote the feature matrix $X=[\boldsymbol{x}^{(1)},\cdots,\boldsymbol{x}^{(n)}]\in\mathbb{R}^{d\times n}$
and the label vector $\boldsymbol{y}=[y_{1},\cdots,y_{n}]{}^{\top}\in\mathbb{R}^{n}$.
The SLM defined in Eq. (\ref{eq:second-order-linear-mode}) can be
written in the matrix form
\begin{equation}
\boldsymbol{y}=X{}^{\top}\boldsymbol{w}^{*}+\mathcal{A}(M^{*})+\boldsymbol{\xi}\label{eq:second-order-linear-model-matrix-form}
\end{equation}
 where the operator $\mathcal{A}(\cdot):\mathbb{R}^{d\times d}\rightarrow\mathbb{R}^{d}$
is defined by $\mathcal{A}(M)\triangleq[\boldsymbol{x}^{(1)}{}^{\top}M\boldsymbol{x}^{(1)},\cdots,\boldsymbol{x}^{(n)}{}^{\top}M\boldsymbol{x}^{(n)}]$.
The operator $\mathcal{A}$ is called the rank-one symmetric matrix
sensing operator since $\boldsymbol{x}{}^{\top}M\boldsymbol{x}=\left\langle \boldsymbol{x}\boldsymbol{x}{}^{\top},M\right\rangle $
where the sensing matrix $\boldsymbol{x}\boldsymbol{x}{}^{\top}$
is of rank-one and symmetric. The adjoint operator of $\mathcal{A}$
is $\mathcal{A}'$ . To make the learning problem well-defined, it
is necessary to assume $M^{*}$ to be a symmetric low-rank matrix
\citep{ming_lin_non-convex_2016}. We assume $\boldsymbol{x}$ is
coordinate sub-gaussian with mean zero and unit variance. The elementwise
third order moment of $\boldsymbol{x}$ is denoted as $\boldsymbol{\kappa}^{*}\triangleq\mathbb{E}\boldsymbol{x}^{3}$
and the fourth order moment is $\boldsymbol{\phi}^{*}\triangleq\mathbb{E}\boldsymbol{x}^{4}$.
For sub-gaussian random variable $z$, we denote its $\psi_{2}$-Orlicz
norm \cite{koltchinskii_oracle_2011} as $\|z\|_{\psi_{2}}$ . Without
loss of generality we assume each coordinate of $\boldsymbol{x}$
is bounded by unit sub-gaussian norm, that is, $\|\boldsymbol{x}_{i}\|_{\psi_{2}}\leq1$
for $i\in\{1,\cdots,d\}$. The matrix Frobenius norm, nuclear norm
and spectral norm are denoted as $\|\cdot\|_{F}$ , $\|\cdot\|_{*}$
, $\|\cdot\|_{2}$ respectively. We use $I$ to denote identity matrix
or identity operator whose dimension or domain can be inferred from
context. $\mathcal{D}(\cdot)$ denotes the diagonal function. For
any two matrices $A$ and $B$, we denote their Hadamard product as
$A\circ B$. The elementwise squared matrix is defined by $A^{2}\triangleq A\circ A$.
For a non-negative real number $\xi\geq0$, the symbol $O(\xi)$ denotes
some perturbation matrix whose spectral norm is upper bounded by $\xi$
. The $i$-th largest singular value of matrix $M$ is $\sigma_{i}(M)$
. To abbreviate our high probability bounds, given a probability $\eta$,
we use symbol $C_{\eta}$ and $c_{\eta}$ to denote some factors polynomial
logarithmic in $1/\eta$ and any other necessary variables that do
not change the polynomial order of our bounds. 

Estimating $\{\boldsymbol{w}^{*},M^{*}\}$ with $n\ll O(d^{2})$ is
an ill-proposed problem without additional structure knowledge about
$M^{*}$. In matrix sensing literatures, the most popular assumption
is that $M^{*}$ is low-rank. Following the standard convex relaxation,
we could penalize the rank of $M$ approximately by nuclear norm which
leads to the convex optimization problem
\begin{equation}
\min_{\boldsymbol{w},M}\ \frac{1}{2n}\|X{}^{\top}\boldsymbol{w}+\mathcal{A}(M)-\boldsymbol{y}\|^{2}+\frac{\lambda}{n}\|M\|_{*}\ .\label{eq:min-nuclear-norm}
\end{equation}
 Although the state-of-the-art nuclear norm solvers can handle large
scale problems when the feature is sparse, minimizing Eq. (\ref{eq:min-nuclear-norm})
is still computationally expensive. An alternative more efficient
approach is to decompose $M$ as product of two low-rank matrices
$M=UV{}^{\top}$ where $U,V\in\mathbb{R}^{d\times k}$. To this end
we turn to a nonconvex optimization problem:
\begin{equation}
\min_{\boldsymbol{w},U,V}\ \mathcal{L}(\boldsymbol{w},U,V)\triangleq\frac{1}{2n}\|X{}^{\top}\boldsymbol{w}+\mathcal{A}(UV{}^{\top})-\boldsymbol{y}\|^{2}\ .\label{eq:non-convex-optimization-loss}
\end{equation}
Heuristically, one can solve Eq. (\ref{eq:non-convex-optimization-loss})
by updating $\boldsymbol{w},U,V$ via alternating gradient descent.
Due to the nonconvexity, it is challenging to derive the global convergent
guarantee for this kind of heuristic algorithms. If the problem is
simple enough, such as the asymmetric matrix sensing problem, the
heuristic alternating gradient descent might work well. However, in
our problem this is no longer true. Naive gradient descent will lead
to non-convergent behavior due to the symmetric matrix sensing. To
design a global convergent nonconvex algorithm, we need a novel gradient-free
learning framework which we call the moment-estimation-sequence method.
We will present the high level idea of this technique in subsection
\ref{sub:Learning-Nonconvex-Model}.

\subsection{Learning without Gradient Descent\label{sub:Learning-Nonconvex-Model}}

In this subsection, we will discuss several fundamental challenges
of learning the SLM. We will show that the conventional gradient descent
is no longer a good heuristic. This motivates us looking for a gradient-free
approach which leads to the moment-estimation-sequence method.

To see why gradient descent is a bad idea, let us compute the expected
gradient of $\mathcal{L}(\boldsymbol{w}^{(t)},U^{(t)},V^{(t)})$ with
respect to $V^{(t)}$ at step $t$. 
\begin{align}
\mathbb{E}\nabla_{V}\mathcal{L}(\boldsymbol{w}^{(t)},U^{(t)},V^{(t)})= & 2(M^{(t)}-M^{*})U^{(t)}+F^{(t)}U^{(t)}\label{eq:expect-gradient-V-loss}
\end{align}
 where $F^{(t)}=\mathrm{tr}(M^{(t)}-M^{*})I+\mathcal{D}(\boldsymbol{\phi}-3)\mathcal{D}(M^{(t)}-M^{*})+\mathcal{D}(\boldsymbol{\kappa})\mathcal{D}(\boldsymbol{w}^{(t)}-\boldsymbol{w}^{*})$.
In previous researches, one expects $\mathbb{E}\nabla\mathcal{L}\approx I$.
However this is no longer the case in our problem. From Eq. (\ref{eq:expect-gradient-V-loss}),
$\|\frac{1}{2}\mathbb{E}\nabla\mathcal{L}-I\|_{2}$ is dominated by
$\|\boldsymbol{\kappa}\|_{\infty}$ and $\|\boldsymbol{\phi}-3\|_{\infty}$
. For non-Gaussian distributions, these two perturbation terms can
be easily large enough to prevent the fast convergence of the algorithm.
The slow convergence not only appears in the theoretical analysis
but also is observable in numerical experiments. Please check our
experiment section for simulation results of gradient descent algorithm
with slow convergence behavior. The gradient of $\boldsymbol{w}$
is similarly biased by $O(\|\boldsymbol{\kappa}\|_{\infty})$.

The failure of gradient descent inspires us looking for a gradient-free
learning method. The perturbation terms in Eq. (\ref{eq:expect-gradient-V-loss})
are high order moments of sub-gaussian variable $\boldsymbol{x}$.
It might be possible to construction a sequence of high order moments
to eliminate these perturbation terms. We call this idea the moment-estimation-sequence
method. 

The next critical question is whether the desired moment estimation
sequence exists and how to construct it efficiently. Unfortunately,
specific to the SLM on sub-gaussian distribution, this is impossible
in general. We need an addition but mild enough assumption on the
sub-gaussian distribution which we call the \emph{Moment Invertible Property }
(MIP).
\begin{defn}[Moment Invertible Property]
\label{def:Moment-Invertible-Property} A sub-gaussian distribution
is called $\tau$-Moment Invertible if $|\phi-1-\kappa^{2}|\geq\tau$
for some constant $\tau>0$.
\end{defn}
The definition of $\tau$-MIP is motivated by our estimation sequence
construction. When the MIP cannot be satisfied, one cannot eliminate
the perturbation terms via the moment-estimation-sequence method and
no global convergence rate to $M^{*}$ can be guaranteed. An exemplar
distribution doesn't satisfy the MIP is the Bernoulli distribution.
In order to learn the SLM on non-MIP distributions, we need to further
assume $M^{*}$ to be diagonal-free. That is, $M^{*}=\bar{M}^{*}-\mathcal{D}(\bar{M}^{*})$
where $\bar{M}^{*}$ is low-rank and symmetric. It is interesting
to note that $M^{*}$ in this case is actually full-rank but still
recoverable since we have the knowledge about its low-rank structure
$\bar{M}^{*}$.

\section{The Moment-Estimation-Sequence Method \label{sec:Moment-Estimation-Sequence}}

In this section, we construct the moment estimation sequence for MIP
distribution in Algorithm \ref{alg:moment-estimation-sequence-method}
and non-MIP distribution in subsection \ref{sub:non-MIP-moment-estimation-sequence}.
We will focus on the high level intuition of our construction in this
section. The theoretical analysis is given in Section \ref{sec:Theoretical-Analysis}. 

\begin{algorithm}
\begin{algorithmic}[1]

\REQUIRE The mini-batch size $n$; number of total update $T$; training
instances $X^{(t)}\triangleq[\boldsymbol{x}^{(t,1)},\boldsymbol{x}^{(t,2)},\cdots,\boldsymbol{x}^{(t,n)}]$,
$\boldsymbol{y}^{(t)}\triangleq[y^{(t,1)},y^{(t,2)},\cdots,y^{(t,n)}]{}^{\top}$;
rank $k\geq1$ .

\ENSURE $\boldsymbol{w}^{(T)},U^{(T)},V^{(T)},M^{(t)}\triangleq U^{(t)}V^{(t)}{}^{\top}$.

\STATE For any $\boldsymbol{z}\in\mathbb{R}^{n}$ and $M\in\mathbb{R}^{d\times d}$,
define function 
\begin{align*}
 & \mathcal{P}^{(t,0)}(\boldsymbol{z})\triangleq\boldsymbol{1}{}^{\top}\boldsymbol{z}/n\quad\mathcal{P}^{(t,1)}(\boldsymbol{z})\triangleq X^{(t)}\boldsymbol{z}/n\quad\mathcal{P}^{(t,2)}(\boldsymbol{z})\triangleq(X^{(t)})^{2}\boldsymbol{z}/n-\mathcal{P}^{(t,0)}(\boldsymbol{z})\\
 & \mathcal{A}^{(t)}(M)\triangleq\mathcal{D}(X^{(t)}{}^{\top}MX^{(t)})\quad\mathcal{H}^{(t)}(\boldsymbol{z})\triangleq\mathcal{A}^{(t)}{}'\mathcal{A}^{(t)}(\boldsymbol{z})/(2n)\ .
\end{align*}

\STATE Retrieve $n$ training instances to estimate the third and
fourth order moments $\boldsymbol{\kappa}$ and $\boldsymbol{\phi}$
.

\STATE For $j\in\{1,\cdots,d\}$, solve $G\in\mathbb{R}^{d\times2}$
and $H\in\mathbb{R}^{d\times2}$ where the $j$-th row of $G$ and
$H$ are
\begin{equation}
G_{i,:}{}^{\top}=\left[\begin{array}{cc}
1 & \boldsymbol{\kappa}_{j}\\
\boldsymbol{\kappa}_{j} & \boldsymbol{\phi}_{j}-1
\end{array}\right]^{-1}\left[\begin{array}{c}
\boldsymbol{\kappa}_{j}\\
\boldsymbol{\phi}_{j}-3
\end{array}\right]\quad H_{i,:}{}^{\top}=\left[\begin{array}{cc}
1 & \boldsymbol{\kappa}_{j}\\
\boldsymbol{\kappa}_{j} & \boldsymbol{\phi}_{j}-1
\end{array}\right]^{-1}\left[\begin{array}{c}
1\\
0
\end{array}\right]\ .\label{eq:solve_g_and_h}
\end{equation}

\STATE Initialize $\boldsymbol{w}^{(0)}=\boldsymbol{0}$, $V^{(0)}=0$.
$U^{(0)}=\mathrm{SVD}(\mathcal{H}^{(0)}(\boldsymbol{y}^{(0)}),k)$,
that is the top-$k$ singular vectors.

\FOR{$t=1,2,\cdots,T$}

\STATE Retrieve $n$ training instances $X^{(t)},\boldsymbol{y}^{(t)}$
, compute
\begin{align*}
 & \hat{\boldsymbol{y}}^{(t)}=X^{(t)}{}^{\top}\boldsymbol{w}^{(t-1)}+\mathcal{A}^{(t)}(U^{(t-1)}V^{(t-1)}{}^{\top})\quad\hat{U}^{(t)}=V^{(t-1)}-\mathcal{M}^{(t)}(\hat{\boldsymbol{y}}^{(t)}-\boldsymbol{y}^{(t)})U^{(t-1)}\\
 & \mathcal{M}^{(t)}(\hat{\boldsymbol{y}}^{(t)}-\boldsymbol{y}^{(t)})\triangleq\mathcal{H}^{(t)}(\hat{\boldsymbol{y}}^{(t)}-\boldsymbol{y}^{(t)})-\frac{1}{2}\mathcal{D}\left(G_{1}\circ\mathcal{P}^{(t,1)}(\hat{\boldsymbol{y}}^{(t)}-\boldsymbol{y}^{(t)})\right)\\
 & \quad\quad\quad\quad\quad\quad\quad\quad\quad-\frac{1}{2}\mathcal{D}\left(G_{2}\circ\mathcal{P}^{(t,2)}(\hat{\boldsymbol{y}}^{(t)}-\boldsymbol{y}^{(t)})\right)
\end{align*}

\STATE Orthogonalize $\hat{U}^{(t)}$ via QR decomposition: $U^{(t)}R^{(t)}=\hat{U}^{(t)}$
, $V^{(t)}=V^{(t-1)}R^{(t)}{}^{\top}$ .

\STATE $\mathcal{W}^{(t)}(\hat{\boldsymbol{y}}^{(t)}-\boldsymbol{y}^{(t)})\triangleq H_{1}\circ\mathcal{P}^{(t,1)}(\hat{\boldsymbol{y}}^{(t)}-\boldsymbol{y}^{(t)})+H_{2}\circ\mathcal{P}^{(t,2)}(\hat{\boldsymbol{y}}^{(t)}-\boldsymbol{y}^{(t)})$
.

\STATE $\boldsymbol{w}^{(t)}=\boldsymbol{w}^{(t-1)}-\mathcal{W}^{(t)}(\hat{\boldsymbol{y}}^{(t)}-\boldsymbol{y}^{(t)})$
.

\ENDFOR

\STATE \textbf{Output:} $\boldsymbol{w}^{(T)},U^{(T)},V^{(T)}$ .\textbf{ }

\end{algorithmic}

\caption{Moment Estimation Sequence Method (MES)}

\label{alg:moment-estimation-sequence-method}
\end{algorithm}

Suppose $\boldsymbol{x}$ is i.i.d. sampled from an MIP distribution.
Our moment estimation sequence is constructed in Algorithm \ref{alg:moment-estimation-sequence-method}.
Denote $\{\boldsymbol{w}^{(t)},M^{(t)}\}$ to be an estimation sequence
of $\{\boldsymbol{w}^{*},M^{*}\}$ where $M^{(t)}=U^{(t)}V^{(t)}{}^{\top}$
. We will show that $\|\boldsymbol{w}^{(t)}-\boldsymbol{w}^{*}\|_{2}+\|M^{(t)}-M^{*}\|_{2}\rightarrow0$
as $t\rightarrow\infty$. The key idea of our construction is to eliminate
$F^{(t)}$ in the expected gradient. By construction, 
\begin{align*}
 & \mathcal{P}^{(t,0)}(\hat{\boldsymbol{y}}^{(t)}-\boldsymbol{y}^{(t)})\approx\mathrm{tr}(M^{(t)}-M^{*})\quad\mathcal{P}^{(t,1)}(\hat{\boldsymbol{y}}^{(t)}-\boldsymbol{y}^{(t)})\approx D(M^{(t)}-M^{*})\boldsymbol{\kappa}+\boldsymbol{w}^{(t)}-\boldsymbol{w}^{*}\\
 & \mathcal{P}^{(t,2)}(\hat{\boldsymbol{y}}^{(t)}-\boldsymbol{y}^{(t)})\approx D(M^{(t)}-M^{*})(\boldsymbol{\phi}-1)+D(\boldsymbol{\kappa})(\boldsymbol{w}^{(t)}-\boldsymbol{w}^{*})\ .
\end{align*}
 This inspires us to find a linear combination of $\mathcal{P}^{(t,\cdot)}$
to eliminate $F^{(t)}$ which leads to the linear equations Eq. (\ref{eq:solve_g_and_h}).
Namely, we want to construct $\{\mathcal{M}^{(t)},\mathcal{W}^{(t)}\}$
such that $\mathcal{M}^{(t)}(\hat{\boldsymbol{y}}^{(t)}-\boldsymbol{y}^{(t)})\approx M^{(t)}-M^{*}$
and $\mathcal{W}^{(t)}(\hat{\boldsymbol{y}}^{(t)}-\boldsymbol{y}^{(t)})\approx\boldsymbol{w}^{(t)}-\boldsymbol{w}^{*}$.
The rows of $G$ are exactly the coefficients  of $\mathcal{P}^{(t,\cdot)}$
we are looking for to construct $\mathcal{M}^{(t)}$. We construct
$\mathcal{W}^{(t)}$ similarly by solving Eq. (\ref{eq:solve_g_and_h}).
In Eq. (\ref{eq:solve_g_and_h}) and (\ref{eq:solve_g_and_h}), the
matrix inversion is numerically stable if and only if the distribution
of $\boldsymbol{x}$ is $\tau$-MIP. For non-MIP distributions, Eq.
(\ref{eq:solve_g_and_h}) is singular therefore we couldn't eliminate
the gradient bias in this case. Please see subsection \ref{sub:non-MIP-moment-estimation-sequence}
for an alternative solution on non-MIP distribution.

The following theorem gives the global convergence rate of Algorithm
\ref{alg:moment-estimation-sequence-method} under noise-free condition.
\begin{thm}
\label{thm:global-convergence-rate-of-Moment-Estimation-Sequence}
In Algorithm \ref{alg:moment-estimation-sequence-method}, suppose
$\{\boldsymbol{x}^{(t,i)},y^{(t,i)}\}$ are i.i.d. sampled from model
(\ref{eq:second-order-linear-mode}). The vector $\boldsymbol{x}^{(t,i)}$
is coordinate sub-gaussian of mean zero and unit variance. Each dimension
of  $\mathbb{P}(\boldsymbol{x})$ is $\tau$-MIP. The noise term $\boldsymbol{\xi}=\boldsymbol{0}$
and $M^{*}$ is a rank-$k$ matrix. Then with probability at least
$1-\eta$, 
\begin{align*}
 & \|\boldsymbol{w}^{(t)}-\boldsymbol{w}^{*}\|_{2}+\|M^{(t)}-M^{*}\|_{2}\leq\delta^{t}(\|\boldsymbol{w}^{*}\|_{2}+\|M^{*}\|_{2})\ ,
\end{align*}
 provided $n\geq C_{\eta}(p+1)^{2}/\delta^{2}\max\{p/\tau^{2},k^{2}d\}$,
$p\triangleq\max\{1,\|\boldsymbol{\kappa}^{*}\|_{\infty},\|\boldsymbol{\phi}^{*}-3\|_{\infty},\|\boldsymbol{\phi}^{*}-1\|_{\infty}\}$
and
\begin{align}
 & \delta\leq(4\sqrt{5}\sigma_{1}^{*}/\sigma_{k}^{*}+3)\sigma_{k}^{*}\left(4\sqrt{5}\sigma_{1}^{*}+3\sigma_{k}^{*}+4\sqrt{5}\|\boldsymbol{w}^{*}\|_{2}^{2}\right)^{-1}\ .\label{eq:delta-upper-bound}
\end{align}

\end{thm}
In Theorem \ref{thm:global-convergence-rate-of-Moment-Estimation-Sequence},
we measure the quality of our estimation by the recovery error $\|\boldsymbol{w}^{(t)}-\boldsymbol{w}^{*}\|_{2}+\|M^{(t)}-M^{*}\|_{2}$
at step $t$. Choosing a small enough number $\delta$, Algorithm
\ref{alg:moment-estimation-sequence-method} converges linearly with
rate $\delta$. A small $\delta$ will require a large $n\approx O(1/\delta^{2})$.
Equivalently speaking, when $n$ is larger than the required sampling
complexity, the convergence rate is around $\delta^{t}\approx O(n^{-t/2})$.
The sampling complexity is on order of $\max\{O(k^{2}d\},O(1/\tau^{2})\}$
. For the Gaussian distribution $\tau=2$ therefore the sampling complexity
is $O(k^{2}d)$ for nearly Gaussian distribution. When $\tau$ is
small, $\mathbb{P}(\boldsymbol{x})$ is nearly non-MIP therefore we
need the non-MIP construction of the moment estimation sequence which
is presented in subsection \ref{sub:non-MIP-moment-estimation-sequence}. 

Theorem \ref{thm:global-convergence-rate-of-Moment-Estimation-Sequence}
only considers the noise-free case. The noisy result is similar to
Theorem \ref{thm:global-convergence-rate-of-Moment-Estimation-Sequence}
under the small noise condition \citep{ming_lin_non-convex_2016}.
Roughly speaking, our estimation will linearly converge to the statistical
error level if the noise is small and $M^{*}$ is nearly low-rank.
We will leave the noisy case to the journal version of this work.

\subsection{Non-MIP Distribution\label{sub:non-MIP-moment-estimation-sequence}}

For non-MIP distributions, it is no longer possible to construct the
moment estimation sequence in the same way as MIP distributions because
Eq. (\ref{eq:solve_g_and_h}) will be singular. The essential difficulty
is due to the $\mathcal{D}(M^{*})$ related bias terms in the gradient.
Therefore for non-MIP distributions, it is necessary to assume $M^{*}$
to be diagonal-free, that is, $\mathcal{D}(M^{*})=\mathcal{D}(\boldsymbol{0})$.
More specifically, we assume that $\bar{M}^{*}$ is a low-rank matrix
and $M^{*}=\bar{M}^{*}-\mathcal{D}(\bar{M}^{*})$. Please note that
$M^{*}$ might be a full-rank matrix now. 

We follow the construction in Algorithm \ref{alg:moment-estimation-sequence-method}.
We replace $M^{(t)}$ in Algorithm \ref{alg:moment-estimation-sequence-method}
with $M^{(t)}=U^{(t)}V^{(t)}{}^{\top}-\mathcal{D}(U^{(t)}V^{(t)}{}^{\top})$.
Since $\mathcal{D}(M^{(t)}-M^{*})=\mathcal{D}(\boldsymbol{0})$, denote
$\boldsymbol{z}^{(t)}\triangleq\hat{\boldsymbol{y}}^{(t)}-\boldsymbol{y}^{(t)}$,
\begin{align*}
 & \mathcal{P}^{(t,1)}(\boldsymbol{z}^{(t)})\approx\boldsymbol{w}^{(t)}-\boldsymbol{w}^{*}\qquad\mathcal{P}^{(t,2)}(\boldsymbol{z}^{(t)})\approx D(\boldsymbol{\kappa})(\boldsymbol{w}^{(t)}-\boldsymbol{w}^{*})\\
 & \mathcal{H}^{(t)}(\boldsymbol{z}^{(t)})\approx(M^{(t)}-M^{*})+\mathcal{D}(\boldsymbol{\kappa})\mathcal{D}(\boldsymbol{w}^{(t)}-\boldsymbol{w}^{*})/2\ .
\end{align*}
 Therefore, we can construct our moment estimation sequence as following:
\begin{align*}
\mathcal{M}^{(t)}(\boldsymbol{z}^{(t)})= & \mathcal{H}^{(t)}(\boldsymbol{z}^{(t)})-\mathcal{D}[\mathcal{P}^{(t,2)}(\boldsymbol{z}^{(t)})/2]\quad\mathcal{W}^{(t)}(\boldsymbol{z}^{(t)})=\mathcal{P}^{(t,1)}(\boldsymbol{z}^{(t)})\ .
\end{align*}
 The rest part is the same as Algorithm \ref{alg:moment-estimation-sequence-method}.

\section{Theoretical Analysis\label{sec:Theoretical-Analysis}}

In this section, we present the proof sketch of Theorem \ref{thm:global-convergence-rate-of-Moment-Estimation-Sequence}.
Details are postponed to the appendix. Define $\beta_{t}\triangleq\|\boldsymbol{w}^{(t)}-\boldsymbol{w}^{*}\|_{2},\ \gamma_{t}\triangleq\|M^{(t)}-M^{*}\|_{2},\ \epsilon_{t}\triangleq\beta_{t}+\gamma_{t}$.
Our essential idea is to construct 
\begin{align}
 & \mathcal{M}^{(t)}(\hat{\boldsymbol{y}}^{(t)}-\boldsymbol{y}^{(t)})=M^{(t-1)}-M^{*}+O(\delta\epsilon_{t-1})\label{eq:mathca-M-delta-M+plus_small-error}\\
 & \mathcal{W}^{(t)}(\hat{\boldsymbol{y}}^{(t)}-\boldsymbol{y}^{(t)})=\boldsymbol{w}^{(t-1)}-\boldsymbol{w}^{*}+O(\delta\epsilon_{t-1})\ .\nonumber 
\end{align}
 for some small $\delta\geq0$. Once we have constructed Eq. (\ref{eq:mathca-M-delta-M+plus_small-error}),
we can apply the noisy power iteration analysis as in \citep{ming_lin_non-convex_2016}.
The global convergence rate immediately follows from Theorem \ref{thm:globa-convergence-when-estimation-sequence-perturbation-holds}
given below.
\begin{thm}[Theorem 1 in \citep{ming_lin_non-convex_2016}]
\label{thm:globa-convergence-when-estimation-sequence-perturbation-holds}
Suppose $\{M^{(t)},\boldsymbol{w}^{(t)}\}$ constructed in Algorithm
\ref{alg:moment-estimation-sequence-method} satisfy Eq. (\ref{eq:mathca-M-delta-M+plus_small-error}).
The noisy term $\boldsymbol{\xi}=0$ and $M^{*}$ is of rank $k$.
Then after $t$ iteration,
\begin{align*}
\|\boldsymbol{w}^{(t)}-\boldsymbol{w}^{*}\|_{2}+\|M^{(t)}-M^{*}\|_{2}\leq & \delta^{t}(\|\boldsymbol{w}^{*}\|_{2}+\|M^{*}\|_{2})\ ,
\end{align*}
 provided $\delta$ satisfying Eq. (\ref{eq:delta-upper-bound}) .
\end{thm}
Theorem \ref{thm:globa-convergence-when-estimation-sequence-perturbation-holds}
shows that the recovery error of the sequence $\{\boldsymbol{w}^{(t)},M^{(t)}\}$
will converges linearly with rate $\delta$ as long as Eq. (\ref{eq:mathca-M-delta-M+plus_small-error})
holds true. The next question is whether Eq. (\ref{eq:mathca-M-delta-M+plus_small-error})
can be satisfied with a small $\delta$. To answer this question,
we will show that $\mathcal{M}^{(t)}$ and $\mathcal{W}^{(t)}$ are
nearly isometric operators with no more than $O(C_{\eta}k^{2}d)$
samples.

In low-rank matrix sensing, the Restricted Isometric Property (RIP)
of sensing operator $\mathcal{A}$ determinates the sampling complexity
of recovery. However in the SLM, $\mathcal{A}$ is a symmetric rank-one
sensing operator therefore the conventional RIP condition is too strong
to hold true. Following \citep{ming_lin_non-convex_2016}, we introduce
a weaker requirement, the Conditionally Independent RIP (CI-RIP) condition.
\begin{defn}[CI-RIP]
\label{def:CI-RIP} Suppose $k\geq1$, $\delta_{k}>0$, $M$ is a
fixed rank $k$ matrix . A sensing operator $\mathcal{A}$ is called
$\delta_{k}$ CI-RIP if for a fixed $M$, $\mathcal{A}$ is sampled
independently such that 
\begin{align*}
 & (1-\delta_{k})\|M\|_{F}^{2}\leq\|\mathcal{A}(M)\|_{2}^{2}\leq(1+\delta_{k})\|M\|_{F}^{2}\ .
\end{align*}

\end{defn}
Comparing to the conventional RIP condition, the CI-RIP only requires
the isometric property to hold on a fixed low-rank matrix rather than
any low-rank matrix. The corresponding price is that $\mathcal{A}$
should be independently sampled from $M^{(t)}$. This can be achieved
by resampling at each iteration. Since our algorithm converges linearly,
the resampling takes logarithmically more samples therefore it will
not affect the order of sampling complexity.

The CI-RIP defined in Definition \ref{def:CI-RIP} concerns about
the concentration of $\mathcal{A}$ around zero. The next theorem
shows that $\mathcal{A}$ in the SLM concentrates around its expectation.
That is, $\mathcal{A}$ is CI-RIP after shifted by its expectation.
The proof can be found in Appendix \ref{sec:Proof-of-Sub-gaussian-shifted-CI-RIP}.
\begin{thm}[Sub-gaussian shifted CI-RIP]
 \label{thm:subgaussian-shifted-CI-RIP}Under the same settings of
Theorem \ref{thm:global-convergence-rate-of-Moment-Estimation-Sequence},
suppose $d\geq(2+\|\boldsymbol{\phi}^{*}-3\|_{\infty})^{2}$ . Fixed
a rank $k$ matrix $M$, with probability at least $1-\eta$, provided
$n\geq c_{\eta}k^{2}d/\delta^{2}$,
\begin{align*}
 & \frac{1}{n}\mathcal{A}'\mathcal{A}(M)=2M+\mathrm{tr}(M)I+\mathcal{D}(\boldsymbol{\phi}^{*}-3)\mathcal{D}(M)+O(\delta\|M\|_{2}).
\end{align*}

\end{thm}
Theorem \ref{thm:subgaussian-shifted-CI-RIP} is one of the main contributions
of this work. Comparing to previous results, mostly Theorem 4 in \citep{ming_lin_non-convex_2016},
we have several fundamental improvements. First it allows sub-gaussian
distribution which requires a more challenging analysis. Secondly,
the sampling complexity is $O(C_{\eta}k^{2}d)$ which is better than
previous $O(C_{\eta}k^{3}d)$ bounds. Recall that the information-theoretical
low bound requires at least $O(C_{\eta}kd)$ complexity. Therefore
our bound is slightly $O(k)$ looser than the lower bound. The key
ingredient of our proof is to apply matrix Bernstein's inequality
with sub-gaussian Hanson-Wright inequality provided in \citep{rudelson_hanson-wright_2013}.
Please check Appendix \ref{sec:Proof-of-Sub-gaussian-shifted-CI-RIP}
for more details.

Based on the shifted CI-RIP condition of operator $\mathcal{A}$,
it is straightforward to prove the following perturbation bounds.
\begin{lem}
\label{lem:concentration-of-mathcal-Pt} Under the same settings of
Theorem \ref{thm:global-convergence-rate-of-Moment-Estimation-Sequence},
for fixed $\boldsymbol{y}=X{}^{\top}\boldsymbol{w}+\mathcal{A}(M)$
, provided $n\geq C_{\eta}k^{2}d/\delta^{2}$ , then with probability
at least $1-\eta$ , 
\begin{align*}
 & \frac{1}{n}\mathcal{A}'(X{}^{\top}\boldsymbol{w})=\mathcal{D}(\boldsymbol{\kappa}^{*})\boldsymbol{w}+O(\delta\|\boldsymbol{w}\|_{2})\\
 & \mathcal{P}^{(0)}(\boldsymbol{y})\triangleq\frac{1}{n}\boldsymbol{1}{}^{\top}\boldsymbol{y}=\mathrm{tr}(M)+O[\delta(\|\boldsymbol{w}\|_{2}+\|M\|_{2})]\\
 & \mathcal{P}^{(1)}(\boldsymbol{y})\triangleq\frac{1}{n}X\boldsymbol{y}=\mathcal{D}(M)\boldsymbol{\kappa}^{*}+\boldsymbol{w}+O[\delta(\|\boldsymbol{w}\|_{2}+\|M\|_{2})]\\
 & \mathcal{P}^{(2)}(\boldsymbol{y})\triangleq\frac{1}{n}X^{2}\boldsymbol{y}-\mathcal{P}^{(0)}(\boldsymbol{y})=\mathcal{D}(M)(\boldsymbol{\phi}^{*}-1)+\mathcal{D}(\boldsymbol{\kappa}^{*})\boldsymbol{w}+O[\delta(\|\boldsymbol{w}\|_{2}+\|M\|_{2})]\ .
\end{align*}

\end{lem}
Lemma \ref{lem:concentration-of-mathcal-Pt} shows that $\mathcal{A}'X{}^{\top}$
and $\mathcal{P}^{(t,\cdot)}$ are all concentrated around their expectations
with no more than $O(C_{\eta}k^{2}d)$ samples. To finish our construction
of the moment estimation sequence, we need to bound the deviation
of $G$ and $H$ from their expectation $G^{*}$ and $H^{*}$ . This
is done in the following lemma. 
\begin{lem}
\label{lem:error-bound-G_j-G_j_*} Suppose $\mathbb{P}(\boldsymbol{x})$
is $\tau$-MIP. Then in Algorithm \ref{alg:moment-estimation-sequence-method},
for any $j\in\{1,\cdots,d\}$,
\begin{align*}
\|G-G^{*}\|_{\infty}\leq & \delta,\ \|H-H^{*}\|_{\infty}\leq\delta\ ,
\end{align*}
 provided $n\geq C_{\eta}(1+\tau^{-1}\sqrt{\|\boldsymbol{\kappa}^{*}\|_{\infty}^{2}+\|\boldsymbol{\phi}^{*}-3\|_{\infty}^{2}})/(\tau\delta^{2})\ .$
\end{lem}
Lemma \ref{lem:error-bound-G_j-G_j_*} shows that $G\approx G^{*}$
as long as $n\geq O(1/\tau^{2})$. Since $G$ is the solution of Eq.
(\ref{eq:solve_g_and_h}), it requires $\mathbb{P}(\boldsymbol{x})$
must be $\tau$-MIP with $\tau>0$. When $\tau=0$, for example on
binary Bernoulli distribution, we must use the construction in subsection
\ref{sub:non-MIP-moment-estimation-sequence} instead. As the non-MIP
moment estimation sequence doesn't invoke the inversion of moment
matrices, the sampling complexity will not depend on $O(1/\tau)$. 

We are now ready to give the condition of Eq. (\ref{eq:mathca-M-delta-M+plus_small-error})
being true.
\begin{lem}
\label{lem:concentration-of-mathcal-M-and-W} Under the same settings
of Theorem \ref{thm:global-convergence-rate-of-Moment-Estimation-Sequence},
with  probability at least $1-\eta$ , Eq. (\ref{eq:mathca-M-delta-M+plus_small-error})
holds true provided 
\[
n\geq C_{\eta}(p+1)^{2}/\delta^{2}\max\{p/\tau^{2},k^{2}d\}
\]
 where $p\triangleq\max\{1,\|\boldsymbol{\kappa}^{*}\|_{\infty},\|\boldsymbol{\phi}^{*}-3\|_{\infty},\|\boldsymbol{\phi}^{*}-1\|_{\infty}\}$
. 
\end{lem}
Lemma \ref{lem:concentration-of-mathcal-M-and-W} shows that the sampling
complexity to guarantee Eq. (\ref{eq:mathca-M-delta-M+plus_small-error})
is bounded by $O(k^{2}d)$ or $O(1/\tau^{2})$, depending on which
one dominates. The proof of Lemma \ref{lem:concentration-of-mathcal-M-and-W}
consists of two steps. First we replace each operator or matrix with
its expectation plus a small perturbation given in Lemma \ref{lem:concentration-of-mathcal-Pt}
and Lemma \ref{lem:error-bound-G_j-G_j_*}. Then Lemma \ref{lem:concentration-of-mathcal-M-and-W}
follows after simplification. Theorem \ref{thm:global-convergence-rate-of-Moment-Estimation-Sequence}
is obtained by combining Lemma \ref{lem:concentration-of-mathcal-M-and-W}
and Theorem \ref{thm:globa-convergence-when-estimation-sequence-perturbation-holds}.

\section{Numerical Simulation}

\begin{figure*}
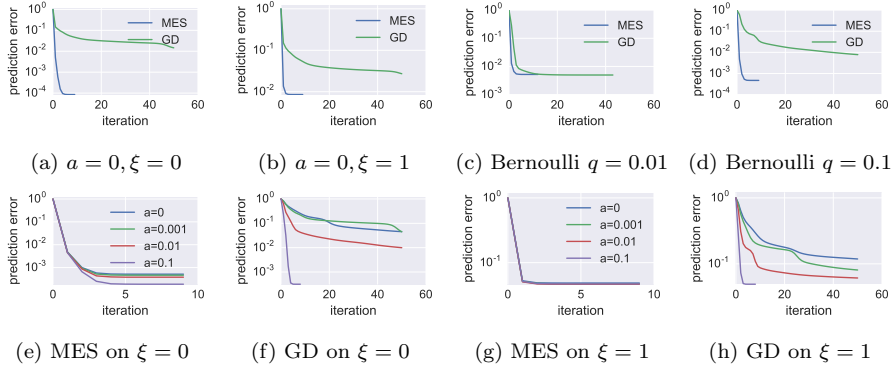

\input{fig_conv_curve_skewGaus_Bern}

\input{fig_conv_curve_Trunc_level}

\caption{Convergence curve. (a)-(b): truncated Gaussian distribution; (c)-(d):
Bernoulli distribution; (e)-(h): different truncation level $a$.}
\label{fig:Convergence-Curve-on-Gaussian-Distribution}
\end{figure*}

In this section, we verify the global convergence rate of Algorithm
\ref{alg:moment-estimation-sequence-method} on MIP and non-MIP distributions.
We will show that the naive gradient descent heuristic cannot work
well when the distribution is skewed. We implement Algorithm \ref{alg:moment-estimation-sequence-method}
in Python. Our computer has 32 GB memory and a 64 bit, 8 core CPU.
Our implementation will be released on our website after publication.
In the following figures, we abbreviate Algorithm \ref{alg:moment-estimation-sequence-method}
as MES and the naive gradient descent as GD.

In the following experiments, we choose the dimension $d=1000$ and
the rank $k=10$. Since non-MIP distributions require $M^{*}$ to
be diagonal-free, we generate $M^{*}$ under two different low-rank
models. For MIP distributions, we randomly generate $U\in\mathbb{R}^{d\times k}$
such that $U{}^{\top}U=I$ . Then we produce $M^{*}=UU{}^{\top}$.
Please note that our model allows symmetric but non-PSD $M^{*}$ but
due to space limitations we only demonstrate the PSD case in this
work. For non-MIP distributions, we generate $\bar{M}^{*}=UU{}^{\top}$
similarly and take $M^{*}=\bar{M}^{*}-\mathcal{D}(\bar{M}^{*})$.
The $\boldsymbol{w}^{*}$ is randomly sampled from $\mathcal{N}(\boldsymbol{0},1/d)$.
The noise term $\boldsymbol{\xi}$ is sampled from $\xi\cdot\mathcal{N}(\boldsymbol{0},I)$
where $\xi$ is the noise level in set $\{0,1\}$. All synthetic experiments
are repeated 10 trials in order to report the average performance.
In each trial, we randomly sample $30kd$ training instances and $10,000$
testing instances. The running time is measured by the number of iterations.
It takes around $7.6$ seconds per iteration on our computer. The
estimation accuracy is measured by the normalized mean squared error
$[\mathbb{E}(y_{\mathrm{pred}}-y_{\mathrm{true}})^{2}]/\mathbb{E}(y_{\mathrm{true}}^{2})$.
We terminate the experiment after $50$ iterations or when the training
error decreases less than $10^{-8}$ between two consecutive iterations.

In Figure \ref{fig:Convergence-Curve-on-Gaussian-Distribution} (a)-(b),
we report the convergence curve on truncated Gaussian distribution.
To sample $\boldsymbol{x}$, we first generate a Gaussian random number
$\hat{x}$ then truncate $x=\min\{\hat{x},a\}$ where $a$ is the
truncation level. In Figure \ref{fig:Convergence-Curve-on-Gaussian-Distribution}
(a)-(b) the truncate level $a=0$. Our method MES converges linearly
and is significantly faster than GD. 

In Figure \ref{fig:Convergence-Curve-on-Gaussian-Distribution} (c)-(d),
we report the convergence curve on Bernoulli distribution which is
non-MIP. We set $\mathcal{D}(M^{*})=0$. We choose the binary Bernoulli
distribution where $\mathbb{P}(x=1)=q$ otherwise $x=0$. In (c) $q=0.01$
and in (d) $q=0.1$. Again MES converges much faster in both (c) and
(d). 

As we analyzed in Section \ref{sec:Theoretical-Analysis}, the failure
of the gradient descent heuristic in the SLM is because the gradient
is bias by $O(\|\boldsymbol{\kappa}\|_{\infty})$. We expect the convergence
of GD being worse when the skewness of the distribution is larger.
To verify this, we report the convergence curves of MES and GD on
truncated  Gaussian with $a=\{0,10^{-3},0.01,0.1\}$ in Figure \ref{fig:Convergence-Curve-on-Gaussian-Distribution}
(e) and (h) under different noise level. As we expected, when $a\rightarrow0$
, the skewness becomes larger and GD converges worse. When $a=0$
, GD is unable to find the global optimal solution at all. In contrast,
MES always converges globally and linearly under any $a$.

\section{Conclusion}

We develop the first provable nonconvex algorithm for learning the
second order linear model with $O(k^{2}d)$ sampling complexity. This
theoretical break-through is built on several recent advances in random
matrix theory such as sub-gaussian Hanson-Wright inequality and our
novel powerful moment-estimation-sequence method. Our analysis reveals
that in high order statistical model, the gradient descent may be
sub-optimal due to the gradient bias induced by the high order moments.
The proposed MES method is the first efficient tool to eliminate such
bias in order to construct a fast convergent sequence for learning
high order linear models. We hope this work could inspire future researches
of nonconvex high order machines.

\small
\bibliographystyle{plainnat}
\bibliography{refs}

\onecolumn
\appendix

\section{Preliminary}

The $\psi_{2}$-Orlicz norm of a random sub-gaussian variable $z$
is defined by
\[
\|z\|_{\psi_{2}}\triangleq\inf\{t>0:\mathbb{E}\exp(z^{2}/t^{2})\leq c\}
\]
 where $c>0$ is a constant. For a random sub-gaussian vector $\boldsymbol{z}\in\mathbb{R}^{n}$,
its $\psi_{2}$-Orlicz norm is 
\[
\|\boldsymbol{z}\|_{\psi_{2}}\triangleq\sup_{\boldsymbol{x}\in S^{n-1}}\|\left\langle \boldsymbol{z},\boldsymbol{x}\right\rangle \|_{\psi_{2}}
\]
 where $S^{n-1}$ is the unit sphere.

The following theorem gives the matrix Bernstein's inequality \cite{roman_vershynin_high-dimensional_2017}. 
\begin{thm}[Matrix Bernstein's inequality]
 \label{thm:matrix-berstein} Let $X_{1},\cdots,X_{N}$ be independent,
mean zero $d\times n$ random matrices with $d\geq n$ and $\|X_{i}\|_{2}\leq B$.
Denote 
\[
\sigma^{2}\triangleq\max\{\|\sum_{i=1}^{N}\mathbb{E}X_{i}X_{i}{}^{\top}\|_{2},\|\sum_{i=1}^{N}\mathbb{E}X_{i}{}^{\top}X_{i}\|_{2}\}\ .
\]
 Then for any $t\geq0$, we have
\[
\mathbb{P}(\|\sum_{i=1}^{N}X_{i}\|_{2}\geq t)\leq2d\exp\left[-c\min\left(\frac{t^{2}}{\sigma^{2}},\frac{t}{B}\right)\right]\ .
\]
 where $c$ is a universal constant. Equivalently, with probability
at least $1-\eta$, 
\[
\|\sum_{i=1}^{N}X_{i}\|_{2}\leq c\max\left\{ B\log(2d/\eta),\sigma\sqrt{\log(2d/\eta)}\right\} \ .
\]
 When $\mathbb{E}X_{i}\not=\boldsymbol{0}$, replacing $X_{i}$ with
$X_{i}-\mathbb{E}X_{i}$ the inequality still holds true.
\end{thm}
The following Hanson-Wright inequality for sub-gaussian variables
is given in \cite{rudelson_hanson-wright_2013} . 
\begin{thm}[Sub-gaussian Hanson-Wright inequality]
 \label{thm:hanson-wright-inequality} Let $\boldsymbol{x}=[x_{1},\cdots,x_{d}]\in\mathbb{R}^{d}$
be a random vector with independent, mean zero, sub-gaussian coordinates.
Then given a fixed $d\times d$ matrix $M$, for any $t\geq0$,
\[
\mathbb{P}\left\{ |\boldsymbol{x}{}^{\top}A\boldsymbol{x}-\mathbb{E}\boldsymbol{x}{}^{\top}A\boldsymbol{x}|\geq t\right\} \leq2\exp\left[-c\min\left(\frac{t^{2}}{B^{4}\|A\|_{F}^{2}},\frac{t}{B^{2}\|A\|_{2}}\right)\right]\ ,
\]
 where $B=\max_{i}\|X_{i}\|_{\psi_{2}}$ and $c$ is a universal positive
constant. Equivalently, with probability at least $1-\eta$, 
\begin{align*}
|\boldsymbol{x}{}^{\top}A\boldsymbol{x}-\mathbb{E}\boldsymbol{x}{}^{\top}A\boldsymbol{x}|\leq & c\max\{B^{2}\|A\|_{2}\log(2/\eta),B^{2}\|A\|_{F}\sqrt{\log(2/\eta)}\}\ .
\end{align*}

\end{thm}
The next lemma estimates the covering number of low-rank matrices
\cite{candes_tight_2011}. 
\begin{lem}[Covering number of low-rank matrices]
 \label{lem:convering-number-of-low-rank-matrices} Let $S=\{M\in\mathbb{R}^{d\times d}:\mathrm{rank}(M)\leq k,\|M\|_{F}\leq c\}$
be the set of rank-$k$ matrices with unit Frobenius norm. Then there
is an $\epsilon$-net cover of $S$, denoted as $\bar{S}(\epsilon)$,
such that
\[
|\bar{S}(\epsilon)|\leq(9c/\epsilon)^{(2d+1)k}\ .
\]

\end{lem}
Note that the original lemma in \cite{candes_tight_2011} bounds $\|M\|_{F}=1$
but in the above lemma we slightly relax to $\|M\|_{F}\leq c$ . The
proof is nearly the same as the original one.

\paragraph{Truncation trick}

As Bernstein's inequality requires boundness of the random variable,
we use the truncation trick in order to apply it on unbounded random
matrices. First we condition on the tail distribution of random matrices
to bound the norm of a fixed random matrix. Then we take union bound
over all $n$ random matrices in the summation. The union bound will
result in an extra $O[\log(n)]$ penalty in the sampling complexity
which can be absorbed into $C_{\eta}$ or $c_{\eta}$ . Please check
\citep{tao_topics_2012} for more details.

\section{Proof of Theorem \ref{thm:subgaussian-shifted-CI-RIP} \label{sec:Proof-of-Sub-gaussian-shifted-CI-RIP}}

Define $p_{1}=2+\|\boldsymbol{\phi}^{*}-3\|_{\infty}$ . Recall that
\begin{align*}
\frac{1}{n}\mathcal{A}'\mathcal{A}(M)= & \frac{1}{n}\sum_{i=1}^{n}\boldsymbol{x}^{(i)}\boldsymbol{x}^{(i)}{}^{\top}M\boldsymbol{x}^{(i)}\boldsymbol{x}^{(i)}{}^{\top}\ .
\end{align*}
 Denote
\begin{align*}
 & Z_{i}\triangleq\boldsymbol{x}^{(i)}\boldsymbol{x}^{(i)}{}^{\top}M\boldsymbol{x}^{(i)}\boldsymbol{x}^{(i)}{}^{\top}\\
 & \mathbb{E}Z_{i}=2M+\mathrm{tr}(M)I+\mathcal{D}(\boldsymbol{\phi}^{*}-3)\mathcal{D}(M)\ .
\end{align*}
 In order to apply matrix Bernstein's inequality , we have
\begin{align*}
\|Z_{i}\|_{2}= & \|\boldsymbol{x}^{(i)}\boldsymbol{x}^{(i)}{}^{\top}M\boldsymbol{x}^{(i)}\boldsymbol{x}^{(i)}{}^{\top}\|_{2}\\
\leq & |\boldsymbol{x}^{(i)}{}^{\top}M\boldsymbol{x}^{(i)}|\|\boldsymbol{x}^{(i)}\boldsymbol{x}^{(i)}{}^{\top}\|_{2}\\
\leq & |\boldsymbol{x}^{(i)}{}^{\top}M\boldsymbol{x}^{(i)}|\|\boldsymbol{x}^{(i)}\|_{2}^{2}\\
\leq & c_{\eta}[\|M\|_{F}+|\mathrm{tr}(M)|]\|\boldsymbol{x}^{(i)}\|_{2}^{2}\\
\leq & c_{\eta}[\|M\|_{F}+|\mathrm{tr}(M)|]d\ .
\end{align*}
 And
\begin{align*}
\|\mathbb{E}Z_{i}\|_{2}= & \|2M+\mathrm{tr}(M)I+\mathcal{D}(\boldsymbol{\phi}^{*}-3)\mathcal{D}(M)\|_{2}\\
\leq & 2\|M\|_{2}+|\mathrm{tr}(M)|+\|\boldsymbol{\phi}^{*}-3\|_{\infty}\|M\|_{2}\\
\leq & (2+\|\boldsymbol{\phi}^{*}-3\|_{\infty})\|M\|_{2}+|\mathrm{tr}(M)|\\
\leq & p_{1}\|M\|_{2}+|\mathrm{tr}(M)|\ .
\end{align*}
 And
\begin{align*}
\|\mathbb{E}Z_{i}Z_{i}{}^{\top}\|_{2}= & \|\mathbb{E}\boldsymbol{x}^{(i)}\boldsymbol{x}^{(i)}{}^{\top}M\boldsymbol{x}^{(i)}\boldsymbol{x}^{(i)}{}^{\top}\boldsymbol{x}^{(i)}\boldsymbol{x}^{(i)}{}^{\top}M\boldsymbol{x}^{(i)}\boldsymbol{x}^{(i)}{}^{\top}\|_{2}\\
\leq & c_{\eta}d\|\mathbb{E}\boldsymbol{x}^{(i)}\boldsymbol{x}^{(i)}{}^{\top}M\boldsymbol{x}^{(i)}\boldsymbol{x}^{(i)}{}^{\top}M\boldsymbol{x}^{(i)}\boldsymbol{x}^{(i)}{}^{\top}\|_{2}\\
\leq & c_{\eta}d\|\mathbb{E}\boldsymbol{x}^{(i)}\boldsymbol{x}^{(i)}{}^{\top}\|_{2}|\boldsymbol{x}^{(i)}{}^{\top}M\boldsymbol{x}^{(i)}|^{2}\\
\leq & c_{\eta}d\|\mathbb{E}\boldsymbol{x}^{(i)}\boldsymbol{x}^{(i)}{}^{\top}\|_{2}[\|M\|_{F}+|\mathrm{tr}(M)|]^{2}\\
\leq & c_{\eta}d[\|M\|_{F}+|\mathrm{tr}(M)|]^{2}\ .
\end{align*}
 And
\begin{align*}
\|(\mathbb{E}Z_{i})(\mathbb{E}Z_{i}){}^{\top}\|_{2}\leq & \|\mathbb{E}Z_{i}\|_{2}^{2}\\
\leq & [p_{1}\|M\|_{2}+|\mathrm{tr}(M)|]^{2}\ .
\end{align*}
 Therefore we get
\begin{align*}
\|Z_{i}-\mathbb{E}Z_{i}\|_{2}\leq & \|Z_{i}\|_{2}+\|\mathbb{E}Z_{i}\|_{2}\\
\leq & c_{\eta}[\|M\|_{F}+|\mathrm{tr}(M)|]d+p_{1}\|M\|_{2}+|\mathrm{tr}(M)|\ .
\end{align*}
 And
\begin{align*}
\mathrm{Var}1\triangleq & \|(Z_{i}-\mathbb{E}Z_{i})(Z_{i}-\mathbb{E}Z_{i}){}^{\top}\|_{2}\\
\leq & \|Z_{i}Z_{i}{}^{\top}\|_{2}+\|(\mathbb{E}Z_{i})(\mathbb{E}Z_{i}){}^{\top}\|_{2}\\
\leq & c_{\eta}d[\|M\|_{F}+|\mathrm{tr}(M)|]^{2}+[p_{1}\|M\|_{2}+|\mathrm{tr}(M)|]^{2}\ .
\end{align*}
 Suppose that
\begin{align*}
 & d[\|M\|_{F}+|\mathrm{tr}(M)|]^{2}\geq[p_{1}\|M\|_{2}+|\mathrm{tr}(M)|]^{2}\\
\Leftarrow & d[\|M\|_{2}+|\mathrm{tr}(M)|]^{2}\geq[p_{1}\|M\|_{2}+|\mathrm{tr}(M)|]^{2}\\
\Leftarrow & d[\|M\|_{2}+|\mathrm{tr}(M)|]^{2}\geq p_{1}^{2}[\|M\|_{2}+|\mathrm{tr}(M)|]^{2}\\
\Leftarrow & d\geq p_{1}^{2}\ .
\end{align*}
 And suppose that 
\begin{align*}
 & [\|M\|_{F}+|\mathrm{tr}(M)|]d\geq p_{1}\|M\|_{2}+|\mathrm{tr}(M)|\\
\Leftarrow & d\geq p_{1}\\
\Leftarrow & d\geq p_{1}^{2}\ .
\end{align*}
The we get
\begin{align*}
\|Z_{i}-\mathbb{E}Z_{i}\|_{2}\leq & c_{\eta}[\|M\|_{F}+|\mathrm{tr}(M)|]d\\
\leq & c_{\eta}kd\|M\|_{2}\\
\mathrm{Var}1\leq & c_{\eta}d[\|M\|_{F}+|\mathrm{tr}(M)|]^{2}\\
\leq & c_{\eta}k^{2}d\|M\|_{2}^{2}\ .
\end{align*}
 Then according to matrix Bernstein's inequality,
\begin{align*}
\|\frac{1}{n}\sum_{i=1}^{n}Z_{i}-\mathbb{E}Z_{i}\|_{2}= & c_{\eta}\max\{\frac{1}{n}kd\|M\|_{2},\frac{1}{\sqrt{n}}\sqrt{k^{2}d}\|M\|_{2}\}\\
\leq & c_{\eta}\frac{1}{\sqrt{n}}\sqrt{k^{2}d}\|M\|_{2}\ .
\end{align*}
 provided
\begin{align*}
 & \frac{1}{n}kd\|M\|_{2}\leq\frac{1}{\sqrt{n}}\sqrt{k^{2}d}\|M\|_{2}\\
\Leftarrow & n\geq d\ .
\end{align*}
 Choose $n\geq c_{\eta}k^{2}d/\delta^{2}$, we getv
\begin{align*}
\|\frac{1}{n}\sum_{i=1}^{n}Z_{i}-\mathbb{E}Z_{i}\|_{2}\leq & \delta\|M\|_{2}\ .
\end{align*}

\section{Proof of Lemma \ref{lem:concentration-of-mathcal-Pt}}
\begin{proof}
To prove $\frac{1}{n}\mathcal{A}'(X{}^{\top}\boldsymbol{w})$,
\begin{align*}
\frac{1}{n}\mathcal{A}'(X{}^{\top}\boldsymbol{w})= & \frac{1}{n}\sum_{i=1}^{n}\boldsymbol{x}^{(i)}\boldsymbol{x}^{(i)}{}^{\top}\boldsymbol{w}\boldsymbol{x}^{(i)}{}^{\top}\ .
\end{align*}
 Similar to Theorem \ref{thm:subgaussian-shifted-CI-RIP}, just replacing
$\mathcal{A}(M)$ with $\boldsymbol{w}$, then with probability at
last $1-\eta$, 
\begin{align*}
\|\frac{1}{n}\mathcal{A}'(X{}^{\top}\boldsymbol{w})-\mathcal{D}(\boldsymbol{\kappa}^{*})\boldsymbol{w}\|_{2}\leq & C_{\eta}\sqrt{d/n}\|\boldsymbol{w}\|_{2}\ .
\end{align*}
 Therefore let
\begin{align*}
 & n\geq C_{\eta}d/\delta^{2}\ .
\end{align*}
 We have
\begin{align*}
 & \|\frac{1}{n}\mathcal{A}'(X{}^{\top}\boldsymbol{w})-\mathcal{D}(\boldsymbol{\kappa}^{*})\boldsymbol{w}\|_{2}\leq\delta\|\boldsymbol{w}\|_{2}\ .
\end{align*}

To prove $\mathcal{P}^{(0)}(\boldsymbol{y})$, 
\begin{align*}
\mathcal{P}^{(0)}(\boldsymbol{y})= & \frac{1}{n}\sum_{i=1}^{n}\boldsymbol{x}^{(i)}{}^{\top}\boldsymbol{w}+\frac{1}{n}\sum_{i=1}^{n}\boldsymbol{x}^{(i)}{}^{\top}M\boldsymbol{x}^{(i)}\ .
\end{align*}
Since $\boldsymbol{x}$ is coordinate sub-gaussian, any $i\in\{1,\cdots,d\}$,
with probability at least $1-\eta$, 
\[
\|\boldsymbol{x}^{(i)}{}^{\top}\boldsymbol{w}\|_{2}\leq c\sqrt{d}\|\boldsymbol{w}\|_{2}\log(n/\eta)\ .
\]
 Then we have
\begin{align*}
\|\frac{1}{n}\sum_{i=1}^{n}\boldsymbol{x}^{(i)}{}^{\top}\boldsymbol{w}-0\|_{2}\leq & C\sqrt{d}\|\boldsymbol{w}\|_{2}\log(n/\eta)/\sqrt{n}\ .
\end{align*}
 Choose $n\geq c_{\eta}d$, we get
\begin{align*}
\|\frac{1}{n}\sum_{i=1}^{n}\boldsymbol{x}^{(i)}{}^{\top}\boldsymbol{w}\|_{2}\leq & \delta\|\boldsymbol{w}\|_{2}\ .
\end{align*}
 From Hanson-Wright inequality,
\begin{align*}
\|\frac{1}{n}\sum_{i=1}^{n}\boldsymbol{x}^{(i)}{}^{\top}M\boldsymbol{x}^{(i)}-\mathrm{tr}(M)\|_{2}\leq & C\|M\|_{F}\log(1/\eta)\\
\leq & C\|M\|_{2}\sqrt{k/n}\log(1/\eta)\ .
\end{align*}
 Therefore
\begin{align*}
\mathcal{P}^{(0)}(\boldsymbol{y})= & \mathrm{tr}(M)+O[(\sqrt{d}\|\boldsymbol{w}\|_{2}+\|M\|_{2}\sqrt{k})/\sqrt{n}\log(n/\eta)]\\
= & \mathrm{tr}(M)+O[C_{\eta}(\sqrt{d}\|\boldsymbol{w}\|_{2}+\|M\|_{2}\sqrt{k})/\sqrt{n}]\\
= & \mathrm{tr}(M)+O[C_{\eta}(\|\boldsymbol{w}\|_{2}+\|M\|_{2}\sqrt{k})\sqrt{d/n}]\ .
\end{align*}
 Let
\begin{align*}
 & n\geq C_{\eta}kd/\delta^{2}\ .
\end{align*}
We have
\begin{align*}
\mathcal{P}^{(0)}(\boldsymbol{y})= & \mathrm{tr}(M)+O[\delta(\|\boldsymbol{w}\|_{2}+\|M\|_{2})]\ .
\end{align*}

To prove $\mathcal{P}^{(1)}(\boldsymbol{y})$,
\begin{align*}
\mathcal{P}^{(1)}(\boldsymbol{y})= & \frac{1}{n}\sum_{i=1}^{n}\boldsymbol{x}^{(i)}\boldsymbol{x}^{(i)}{}^{\top}\boldsymbol{w}+\frac{1}{n}\sum_{i=1}^{n}\boldsymbol{x}^{(i)}\boldsymbol{x}^{(i)}{}^{\top}M\boldsymbol{x}^{(i)}\ .
\end{align*}
 From covariance concentration inequality,
\begin{align*}
\|\frac{1}{n}\sum_{i=1}^{n}\boldsymbol{x}^{(i)}\boldsymbol{x}^{(i)}{}^{\top}\boldsymbol{w}-\boldsymbol{w}\|_{2}\leq & c\sqrt{d/n}\|\boldsymbol{w}\|_{2}\log(d/\eta)\\
\leq & C_{\eta}\sqrt{d/n}\|\boldsymbol{w}\|_{2}\ .
\end{align*}
 To bound the second term in $\mathcal{P}^{(1)}(\boldsymbol{y})$,
apply Hanson-Wright inequality again,
\begin{align*}
\|\boldsymbol{x}^{(i)}\boldsymbol{x}^{(i)}{}^{\top}M\boldsymbol{x}^{(i)}\|_{2}\leq & \|\boldsymbol{x}^{(i)}\|_{2}\|\boldsymbol{x}^{(i)}{}^{\top}M\boldsymbol{x}^{(i)}\|_{2}\\
\leq & c[\|M\|_{F}+\mathrm{tr}(M)]\sqrt{d}\log^{2}(nd/\eta)\\
\leq & C_{\eta}k\|M\|_{2}\sqrt{d}\ .
\end{align*}
 By matrix Chernoff's inequality, choose $n\geq c_{\eta}k^{2}d/\delta^{2}$,
\begin{align*}
\|\frac{1}{n}\sum_{i=1}^{n}\boldsymbol{x}^{(i)}\boldsymbol{x}^{(i)}{}^{\top}M\boldsymbol{x}^{(i)}-\mathcal{D}(M)\boldsymbol{\kappa}^{*}\|_{2}\leq & C_{\eta}k\|M\|_{2}\sqrt{d/n}\\
\leq & \delta\|M\|_{2}\ .
\end{align*}
 Therefore we have
\begin{align*}
\mathcal{P}^{(1)}(\boldsymbol{y})=\boldsymbol{w}+\mathcal{D}(M)\boldsymbol{\kappa}^{*}+O & [\delta(\|\boldsymbol{w}\|_{2}+\|M\|_{2})]\ .
\end{align*}

To bound $\mathcal{P}^{(2)}(\boldsymbol{y})$ , first note that 
\begin{align*}
\mathcal{P}^{(2)}(\boldsymbol{y})= & \frac{1}{n}\sum_{i=1}^{n}\boldsymbol{x}^{(i)2}\boldsymbol{x}^{(i)}{}^{\top}\boldsymbol{w}+\frac{1}{n}\sum_{i=1}^{n}\boldsymbol{x}^{(i)2}\boldsymbol{x}^{(i)}{}^{\top}M\boldsymbol{x}^{(i)}-P^{(0)}(\boldsymbol{y})\\
= & \frac{1}{n}\sum_{i=1}^{n}\mathcal{D}(\boldsymbol{x}^{(i)}\boldsymbol{x}^{(i)}{}^{\top}\boldsymbol{w}\boldsymbol{x}^{(i)})+\frac{1}{n}\sum_{i=1}^{n}\mathcal{D}(\boldsymbol{x}^{(i)}\boldsymbol{x}^{(i)}{}^{\top}M\boldsymbol{x}^{(i)}\boldsymbol{x}^{(i)}{}^{\top})-P^{(0)}(\boldsymbol{y})\ .
\end{align*}
 Then similarly,
\begin{align*}
\|\frac{1}{n}\sum_{i=1}^{n}\boldsymbol{x}^{(i)2}\boldsymbol{x}^{(i)}{}^{\top}\boldsymbol{w}-\mathcal{D}(\boldsymbol{\kappa}^{*})\boldsymbol{w}\|_{2}\leq & C_{\eta}\sqrt{d/n}\|\boldsymbol{w}\|_{2}
\end{align*}
\begin{align*}
\|\frac{1}{n}\sum_{i=1}^{n}\boldsymbol{x}^{(i)2}\boldsymbol{x}^{(i)}{}^{\top}M\boldsymbol{x}^{(i)}-\mathrm{tr}(M)-\mathcal{D}(M)(\boldsymbol{\phi}^{*}-1)\|_{2}\leq & C_{\eta}k\|M\|_{2}\sqrt{d/n}\ .
\end{align*}
The last inequality is because Theorem \ref{thm:subgaussian-shifted-CI-RIP}.
Combine all together, choose $n\geq c_{\eta}k^{2}d$, 
\begin{align*}
\mathcal{P}^{(2)}(\boldsymbol{y})= & \mathcal{D}(\boldsymbol{\kappa}^{*})\boldsymbol{w}+\mathcal{D}(M)(\boldsymbol{\phi}^{*}-1)+O(C_{\eta}\sqrt{d/n}\|\boldsymbol{w}\|_{2})\\
 & +O(\|M\|_{2}k\sqrt{d/n})+O[C_{\eta}(\|\boldsymbol{w}\|_{2}+k\|M\|_{2})\sqrt{d/n}]\\
= & \mathcal{D}(\boldsymbol{\kappa}^{*})\boldsymbol{w}+\mathcal{D}(M)(\boldsymbol{\phi}^{*}-1)+O[C_{\eta}(\|\boldsymbol{w}\|_{2}+\|M\|_{2})k\sqrt{d/n}]\\
= & \mathcal{D}(\boldsymbol{\kappa}^{*})\boldsymbol{w}+\mathcal{D}(M)(\boldsymbol{\phi}^{*}-1)+O[\delta(\|\boldsymbol{w}\|_{2}+\|M\|_{2})]\ .
\end{align*}
 
\end{proof}

\section{Proof of Lemma \ref{lem:error-bound-G_j-G_j_*}}

The next lemma bounds the estimation accuracy of $\boldsymbol{\kappa}^{*},\boldsymbol{\phi}^{*}$
. It directly follows sub-gaussian concentration inequality and union
bound.
\begin{lem}
\label{lem:estimation-accuracy-of-3-4-moments} Given $n$ i.i.d.
sampled $\boldsymbol{x}^{(i)}$, $i\in\{1,\cdots,n\}$. With a probability
at least $1-\eta$, 
\begin{align*}
\|\boldsymbol{\kappa}-\boldsymbol{\kappa}^{*}\|_{\infty}\leq & C_{\eta}/\sqrt{n}\\
\|\boldsymbol{\phi}-\boldsymbol{\phi}^{*}\|_{\infty}\leq & C_{\eta}/\sqrt{n}
\end{align*}
 provided $n\geq C_{\eta}d$ . 
\end{lem}
Denote $G^{*}$ as $G$ in Eq. (\ref{eq:solve_g_and_h}) but computed
with $\boldsymbol{\kappa}^{*},\boldsymbol{\phi}^{*}$. The next lemma
bounds $\|G_{j,:}-G_{j,:}^{*}\|_{2}$ for any $j\in\{1,\cdots,d\}$.
\begin{proof}
Denote $\boldsymbol{g}=G_{j}$, $\boldsymbol{g}^{*}=G_{j}^{*}$, $\kappa=\boldsymbol{\kappa}_{j}$,
$\mathbf{\phi=\boldsymbol{\phi}}_{j}$, 
\begin{align*}
A= & \left[\begin{array}{cc}
1 & \boldsymbol{\kappa}_{j}\\
\boldsymbol{\kappa}_{j} & \boldsymbol{\phi}_{j}-1
\end{array}\right],\ \boldsymbol{b}=\left[\begin{array}{c}
\boldsymbol{\kappa}_{j}\\
\boldsymbol{\phi}_{j}-3
\end{array}\right]\\
A^{*}= & \left[\begin{array}{cc}
1 & \boldsymbol{\kappa}_{j}^{*}\\
\boldsymbol{\kappa}_{j}^{*} & \boldsymbol{\phi}_{j}^{*}-1
\end{array}\right],\ \boldsymbol{b}^{*}=\left[\begin{array}{c}
\boldsymbol{\kappa}_{j}^{*}\\
\boldsymbol{\phi}_{j}^{*}-3
\end{array}\right]\ .
\end{align*}
Then $\boldsymbol{g}=A^{-1}\boldsymbol{b}$, $\boldsymbol{g}^{*}=A^{*-1}\boldsymbol{b}^{*}$
. Since $\mathbb{P}(\boldsymbol{x})$ is $\tau$-MIP, $\|A^{*-1}\|_{2}\leq1/\tau$
. From Lemma \ref{lem:estimation-accuracy-of-3-4-moments},
\begin{align*}
\|A-A^{*}\|_{2}\leq & C\log(d/\eta)/\sqrt{n}\\
\|\boldsymbol{b}-\boldsymbol{b}^{*}\|_{2}\leq & C\log(d/\eta)/\sqrt{n}\ .
\end{align*}
 Define $\Delta_{A}\triangleq A-A^{*}$, $\Delta_{b}\triangleq\boldsymbol{b}-\boldsymbol{b}^{*}$,
$\Delta_{g}\triangleq\boldsymbol{g}-\boldsymbol{g}^{*}$, 
\begin{align*}
 & A\boldsymbol{g}=\boldsymbol{b}\\
\Leftrightarrow & (A^{*}+\Delta_{A})(\boldsymbol{g}^{*}+\Delta_{g})=\boldsymbol{b}^{*}+\Delta_{b}\\
\Leftrightarrow & A^{*}\Delta_{g}+\Delta_{A}\boldsymbol{g}^{*}+\Delta_{A}\Delta_{g}=\Delta_{b}\\
\Leftrightarrow & (A^{*}+\Delta_{A})\Delta_{g}=\Delta_{b}-\Delta_{A}\boldsymbol{g}^{*}\\
\Rightarrow & \|(A^{*}+\Delta_{A})\Delta_{g}\|_{2}=\|\Delta_{b}-\Delta_{A}\boldsymbol{g}^{*}\|_{2}\\
\Rightarrow & \|(A^{*}+\Delta_{A})\Delta_{g}\|_{2}\leq\|\Delta_{b}\|_{2}+\|\Delta_{A}\boldsymbol{g}^{*}\|_{2}\\
\Rightarrow & \|(A^{*}+\Delta_{A})\Delta_{g}\|_{2}\leq C\log(d/\eta)/\sqrt{n}+C\log(d/\eta)/\sqrt{n}\|\boldsymbol{g}^{*}\|_{2}\\
\Rightarrow & \|(A^{*}+\Delta_{A})\Delta_{g}\|_{2}\leq C\log(d/\eta)/\sqrt{n}(1+\|\boldsymbol{g}^{*}\|_{2})\\
\Rightarrow & \|(A^{*}+\Delta_{A})\Delta_{g}\|_{2}\leq C\log(d/\eta)/\sqrt{n}(1+\frac{1}{\tau}\|\boldsymbol{b}^{*}\|_{2})\\
\Rightarrow & \|(A^{*}+\Delta_{A})\Delta_{g}\|_{2}\leq C\log(d/\eta)/\sqrt{n}(1+\frac{1}{\tau}\sqrt{\kappa^{2}+(\phi-3)^{2}})\\
\Rightarrow & [\tau-C\log(d)/\sqrt{n}]\|\Delta_{g}\|_{2}\leq C\log(d/\eta)/\sqrt{n}(1+\frac{1}{\tau}\sqrt{\kappa^{2}+(\phi-3)^{2}})\ .
\end{align*}
 When 
\begin{align*}
 & \tau-C\log(d/\eta)/\sqrt{n}\geq\frac{1}{2}\tau\\
\Leftrightarrow & n\geq4C^{2}\log^{2}(d/\eta)/\tau^{2}\ ,
\end{align*}
 we have
\begin{align*}
\|\Delta_{g}\|_{2}\leq & \frac{2C}{\tau\sqrt{n}}\log(d/\eta)(1+\frac{1}{\tau}\sqrt{\kappa^{2}+(\phi-3)^{2}})\ .
\end{align*}
 Since $\Delta_{g}$ is a vector of dimension 2, its $\ell_{2}$-norm
bound also controls its $\ell_{\infty}$-norm bound up to constant.
Choose
\begin{align*}
 & n\geq C_{\eta}\frac{1}{\tau}(1+\frac{1}{\tau}\sqrt{\kappa^{2}+(\phi-3)^{2}})/\delta^{2}\ .
\end{align*}
 We have
\begin{align*}
\|\Delta_{g}\|_{\infty}\leq & \delta\ .
\end{align*}

The proof of $H$ is similar.
\end{proof}

\section{Proof of Lemma \ref{lem:concentration-of-mathcal-M-and-W}}
\begin{proof}
To abbreviate the notation, we omit $\hat{\boldsymbol{y}}^{(t)}-\boldsymbol{y}^{(t)}$
and superscript $t$ in the following proof. Denote $\mathcal{H}^{*}=\mathbb{E}\mathcal{H}$
and the expectation of other operators similarly. By construction
in Algorithm \ref{thm:global-convergence-rate-of-Moment-Estimation-Sequence},
\begin{align*}
\mathcal{M}\triangleq & \mathcal{H}-\frac{1}{2}\mathcal{D}(G_{1}\circ\mathcal{P}^{(1)})-\frac{1}{2}\mathcal{D}(G_{2}\circ\mathcal{P}^{(2)})\\
= & \mathcal{H}^{*}+O[\delta(\alpha_{t-1}+\beta_{t-1})]\\
 & -\frac{1}{2}\mathcal{D}(G_{1}^{*}\circ\mathcal{P}^{*(1)})-\frac{1}{2}\mathcal{D}(G_{2}^{*}\circ\mathcal{P}^{*(2)})\\
 & +O[\|G-G^{*}\|_{\infty}(\|\mathcal{P}^{*(1)}\|_{2}+\|\mathcal{P}^{*(2)}\|_{2})]\\
 & +O[\|G-G^{*}\|_{\infty}\delta(\alpha_{t-1}+\beta_{t-1})]\\
= & M^{(t)}-M^{*}+O[\delta(\alpha_{t-1}+\beta_{t-1})]\\
 & +O[\delta(\|\mathcal{P}^{*(1)}\|_{2}+\|\mathcal{P}^{*(2)}\|_{2})]\\
 & +O[\delta^{2}(\alpha_{t-1}+\beta_{t-1})]\\
= & M^{(t)}-M^{*}+O[\delta(\alpha_{t-1}+\beta_{t-1})]\\
 & +O[\delta(\|\mathcal{P}^{*(1)}\|_{2}+\|\mathcal{P}^{*(2)}\|_{2})]\\
= & M^{(t)}-M^{*}+O[\delta(\alpha_{t-1}+\beta_{t-1})]\\
 & +O[\delta(\alpha_{t-1}\|\boldsymbol{\kappa}^{*}\|_{\infty}+\beta_{t-1})\\
 & +\alpha_{t-1}\|\boldsymbol{\phi}^{*}-1\|_{\infty}+\beta_{t-1}\|\boldsymbol{\kappa}^{*}\|_{\infty}]\\
= & M^{(t)}-M^{*}+O[\delta(\alpha_{t-1}+\beta_{t-1})]+O[\delta p(\alpha_{t-1}+\beta_{t-1})]\\
= & M^{(t)}-M^{*}+O[\delta(p+1)(\alpha_{t-1}+\beta_{t-1})]\ .
\end{align*}
 The above requires
\begin{align*}
n\geq & \max\{C_{\eta}\frac{1}{\tau}(1+\frac{1}{\tau}\sqrt{\kappa^{2}+(\phi-3)^{2}})/\delta^{2},C_{\eta}k^{2}d\}\\
= & \max\{C_{\eta}p(\tau\delta)^{-2},C_{\eta}k^{2}d\}\ .
\end{align*}
 Replace $\delta(p+1)$ with $\delta$, the proof is completed.

To bound $\mathcal{W}^{(t)}(\hat{\boldsymbol{y}}^{(t)}-\boldsymbol{y}^{(t)})$,
similarly we have
\begin{align*}
\mathcal{W}= & G_{1}\circ\mathcal{P}^{(1)}+G_{2}\circ\mathcal{P}^{(2)}\\
= & G_{1}^{*}\circ\mathcal{P}^{*(1)}+G_{2}^{*}\circ\mathcal{P}^{*(2)}\\
 & +O[\|G-G^{*}\|_{\infty}\delta(\alpha_{t-1}+\beta_{t-1})]\\
 & +O[\|G-G^{*}\|_{\infty}(\|\mathcal{P}^{*(1)}\|_{2}+\|\mathcal{P}^{*(2)}\|_{2})]\\
= & \boldsymbol{w}^{(t-1)}-\boldsymbol{w}^{*}+O[\delta^{2}(\alpha_{t-1}+\beta_{t-1})]\\
 & +O[\delta p(\alpha_{t-1}+\beta_{t-1})]\\
= & \boldsymbol{w}^{(t-1)}-\boldsymbol{w}^{*}+O[\delta(p+1)(\alpha_{t-1}+\beta_{t-1})]\ .
\end{align*}
\end{proof}

\end{document}